\documentclass[10pt,twocolumn,letterpaper]{article}
\usepackage[nodayofweek,level]{datetime}
\usepackage{wacv}
\usepackage{times}
\usepackage{epsfig}
\usepackage{graphicx}
\usepackage{amsmath}
\usepackage{amssymb}
\usepackage{url}
\usepackage{wrapfig}
\usepackage{breqn}
\usepackage{algorithm2e}
\usepackage[noend]{algpseudocode}
\usepackage{array,ragged2e}
\usepackage[caption=false]{subfig}
\usepackage{textcomp}
\usepackage{gensymb} 
\usepackage{tabularx,calc}

\wacvfinalcopy 

\def\wacvPaperID{646} 

\ifwacvfinal\fi
\begin{document}

\newenvironment{conditions}
  {\par\vspace{\abovedisplayskip}\noindent
   \tabularx{\columnwidth}{>{}l<{} @{${}={}$} >{\raggedright\arraybackslash}X}}
  {\endtabularx\par\vspace{\belowdisplayskip}}
  
\title{Content Based Image Retrieval from AWiFS Images Repository of IRS Resourcesat-2 Satellite Based on Water Bodies and Burnt Areas}

\author{Suraj Kothawade \\
IIT Bombay, India\\
{\tt\small surajkothawade@cse.iitb.ac.in}
\and
Kunjan Mhaske \\
Rochester Institute of Technology, NY, USA\\
{\tt\small km1556@rit.edu}
\and
Furkhan Shaikh \\
SGGSIE\&T, Nanded, India\\
{\tt\small shaikhmohammed@sggs.ac.in}
\and
Sahil Sharma \\
SGGSIE\&T, Nanded, India\\
{\tt\small sharmasahil@sggs.ac.in}
}

\maketitle
\ifwacvfinal\thispagestyle{empty}\fi

\begin{abstract}
  Satellite Remote Sensing Technology is becoming a major milestone in the prediction of weather anomalies, natural disasters as well as finding alternative resources in proximity using multiple multi-spectral sensors emitting electromagnetic waves at distinct wavelengths. Hence, it is imperative to extract water bodies and burnt areas from orthorectified tiles and correspondingly rank them using similarity measures. Different objects in all the spheres of the earth have the inherent capability of absorbing electromagnetic waves of distant wavelengths. This creates various unique masks in terms of reflectance on the receptor. We propose Dynamic Semantic Segmentation (DSS) algorithms that utilized the mentioned capability to extract and rank Advanced Wide Field Sensor (AWiFS) images according to various features. This system stores data intelligently in the form of a sparse feature vector which drastically mitigates the computational and spatial costs incurred for further analysis. The compressed source image is divided into chunks and stored in the database for quicker retrieval. This work is intended to utilize readily available and cost effective resources like AWiFS dataset instead of depending on advanced technologies like Moderate Resolution Imaging Spectroradiometer (MODIS) for data which is scarce.
\end{abstract}
\section{Introduction}

Everyday, we have new images captured by remote sensing systems and being added to their exploding repositories. These images are often large in size and come from different sensors and hence have different properties. The storage of these images in database along with their features is a challenging task.

Given that satellite images are many folds in resolution than commonplace images and are generated in monolithic amounts, the need of the hour is to devise a system that can analyze these satellite images effectively. The system must be able to process numerous user-queries as well as rank the  images based on their features using similarity measures.

We propose a Dynamic Semantic Segmentation algorithm that gives solutions to two major problems: water-bodies detection and burnt-area detection using AWiFS image repositories. The proposed system makes efficient usage of computational resources using sparse features and drastically reduces storage costs. Moreover, the architecture of the system incorporates automated storage and retrieval of the images by dividing the images into chunks of small size. This helps us in making use of parallel processing along with distributed storage of data.

\subsection{Current State of Art}
\label{Current State of Art}

Traditional methods in image processing like Normalized Difference Water Index (NDWI) and Modified Normalized Difference Water Index (MNDWI) are being used to classify the water bodies from rest of the vegetation areas. These methods are also useful for water resource management using Advanced Wide Field Sensor (AWiFS) dataset that detects and extracts the water bodies from images. In these methods, the thresholds of NDWI and MNDWI along with the Brightness (BRT) combinations are used to distill out the water bodies. Moreover, the Normalized Difference Vegetation Index (NDVI) is used to filter out the atmospheric errors and irrelevant land areas which gives efficient outcomes. \cite{subramaniam2011automated}. 
	
Subramaniam et al. have developed a decision tree like algorithm with empirically set thresholds to extract the water bodies from AWiFS sensor dataset. This system is focused on the reflectance values of separate spectral bands as well as the traditional methods like NDVI and MNDVI (Modified NDVI). Each threshold range is obtained from the graphical mapping of arbitrary pixel values from images of each bands.
However, the exact range of reflectance values were needed to be found for new satellite IRS-ResourceSat-2 as the work of Subramaniam et al. is based on the satellite IRS-ResourceSat-1. 

The research work done by Upreti is mainly focused on color, texture and shape of clouds. The source of data for creating these features is based on a Very High Resolution Radiometer (VHRR) sensor of Indian meteorological satellite, (Kalpana-1) which provides three bands, visible band, thermal infrared band and water vapour band, along with another image dataset obtained from Meteorological \& Oceanographic Satellite Data Archival Centre (MOSDAC) \cite{upreti2011content}. The system's accuracy is based on it's feature extraction module. Cloud detection and their identification are based on their appearance in the respective bands of sensor. For instance, in infrared images, higher level clouds appear brighter than middle level and lower level ones. The gray color level is examined using color histogram, spatial relationship between the pixels of images with different offsets are considered in texture feature, and contour (boundary) based and region (interior) based shape descriptor are considered in shape feature. Also, the similarity computation and ranking based on weighted similarity values are done using euclidean distance. This system uses multiple datasets to obtain the desired features vectors which utilizes the resources and computational power. This can be optimized by using alternative and optimized approach to use a single AWiFS dataset with multiple threshold limits and combination of bands images. Further, the system gives substantial results as well as retrieves images from the database. However, for desired results, every feature vector needs to be iterated to find the similarity. If input is a large amount of query images to compute then there is chance of computational delay. This delays could be avoided if the system has been given a reference of already stored feature vectors while processing the query image.

For spatial extent and distribution analysis of forest fires, Reddy et al. work is based on AWiFS sensor of IRS P6 satellite (Resourcesat-I satellite). Their system carries out comparative analysis with respect to various forest types, elevated regions to demarcate and identify the post burnt area. The Planck's black body radiance shifts to short wavelengths as temperature increases. Hence, SWIR band has the primary role in development of this method. The method uses the Short Wave Infrared (SWIR) band images to detect the fire flumes and burnt areas. The supervised classification technique using maximum likelihood is applied on IRS P6 AWiFS Imagery with appropriate signatures and training sets for forest burnt areas mapping and to classify the land cover categories. The system gives the results of fire occurrence analysis which is carried out to understand the fire frequency from the forest's edge to the interior and disturbance causing sources. However, this research uses the False Color Composite (FCC) images to observe and detect the burnt scars color pattern from rest of the region. The extra processing to obtain FCC images from fusion of multiple bands which increases the resource utilization of system. Also, storage system should be used to store and analyze the previously processed data.

\section{The ResourceSat-II satellite} 
The ResourceSat-2 satellite by Indian Space Research Organization (ISRO), India, is available to utilize its sensor's datasets. It is enhanced in various specifications over it's predecessor Resourcesat-1. The satellite provides  multi-spectral and panchromatic imagery of Earth's surface using the LISS-III (Linear Imaging Self-Scanning Sensor), LISS-IV, and AWiFS (Advanced Wide Field Sensor) which are three optical remote sensors mounted on it \cite{Resourcesat-2_Handbook} \cite{ISRO}.

LISS III gives 23.5m resolution with data quantization of 10 bit in 4 spectral bands having 140Km swath area. It has B2(Green), B3(Red), B4(NIR) and B5(SWIR) bands.

LISS IV gives 5.8m resolution with data quantization of 10 bit in 3 spectral bands having 23.5Km and 70Km swath area. It has bands B2(Green), B3(Red) and B4(NIR). It also has 2 modes namely Multi-Spectral (Mx) mode – 23.5Km swath and Mono mode – 70Km swath.

AWiFS gives resolution of 56m with data quantization of 12 bit in 4 spectral bands having 740Km swath area. It has same bands as LISS III with wavelengths of B2(0.52-0.59 microns), B3(0.62-0.68 microns), B4(0.77-0.86 microns), B5(1.55 – 1.70 microns). It is specialized to cover the wide field with a minimum geometric distortion.

\subsection{The Advanced Wide Field Sensor (AWiFS)} 
The multi-spectral optical remote sensor, AWiFS, is realized in two electro-optic modules viz. AWiFS-A and AWiFS-B, which are tilted by 11.94\degree with respect to nadir, provides a spatial resolution of 56m and a combined swath area of 737 Km. The AWiFS sensor provides 12 bit radiometry through Multi Linear Gain (MLG) technique. The sensor has electro-optic module which contains refractive imaging optics along with pass interference filter, a thermal filter and a 6000 pixels linear array Charge Coupled Devices (CCD) detector for separate bands. The CCDs used in AWiFS and LISS-III are identical. The output signals from each CCD are amplified and digitized into 12 bit data in the video processing electronics. Data is transmitted in 10 bits and restored to 12 bits on ground to utilize the Resourcesat-1 signal channel. The communication mode of this satellite is real-time with ground station or record and playback mode using on-board 480GB capacity Solid State Recorder (SSR). The on-board calibration scheme is through Light Emitting Diodes (LEDs) as in Resourcesat-1. For the visible NIR bands (B2, B3, B4), the calibration is a progressively increasing sequence of 16 intensity levels through exposure control system. SWIR band has calibration sequence through a repetitive cycle of 2048 scan lines \cite{Resourcesat-2_Handbook} \cite{ISRO}. 

\subsection{Experimentation Datasets}
The AWiFS dataset provided by BHUVAN web portal \cite{BHUVAN} of National Remote Sensing Center (NRSC), India, has radiometric calibrated images of each orthorectified tile according to their captured dates. Earth's surface area is mapped into orthorectified tiles according to the swath area and spatial resolution of a particular satellite. Every tile has unique ID which represents the specific area on the map. The meta-data of respective tiles like the date of capture, longitude and latitude are provided by NRSC. This data is mainly used in image pre-processing calculations. These images have raster data saved in \textit{.tif} format which has a resolution of 2264 x 2264 pixels with a 16 bit depth. These pixel values are called as a digital numbers (DN). Each orthorectified tile has a distinct image for each band (B2, B3, B4, B5) which can be combined or used individually as an input to our DSS algorithm \ref{DSS}. Each image has single channel data and the alternation in color depth could cause the loss of relevant pixel data in gray-scale shades. 

\section{Preprocessing}
The perfectly measured and calibrated dataset is crucial while performing image processing algorithms. Obligated to NRSC's Bhuvan web portal, the dataset available on the site is orthorectified as well as radiometric calibrated. The satellite sensors capture the raw images which are raster images of the captured area. Those values take into account all the radiance received by the sensor. Earth's surface as well as atmospheric irradiance energy creates a haze like mask in images. The purpose of the preprocessing is to create the enhanced images with proper visibility of region of interest before further processing. The thresholds of DSS algorithm depend on these corrected values. It mainly contains removal of solar radiance, dust, smoke particles illumination and provide the top of atmosphere reflectance values over the radiance values from satellite sensors. Reflectance is the ratio and hence a unit less entity which can be efficiently used to perform arithmetic operations.

\subsection{Spectral Radiance}
The radiance is flux of energy (irradiance energy) per solid angle which leaves at a unit surface area in a given direction. Radiance is partialy dependent on reflectance. The spectral radiance is the radiance of a surface per unit frequency or wavelength. In this case, the spectrum is taken as function of wavelength. Hence the calculation are carried out using the same. The spectral radiance at sensor's aperture is measured in watts per steradian per square meter(W·sr$^{-1}$·m$^{-2}$) with respect to the wavelength ($\mu$m) of each band \cite{RadianceUSGS}.

To calculate the spectral radiance, the meta-data of sensors as well as images are necessary. This contains Quantized Calibrated Pixel(QCAL) values known as Digital Number (DN) which are spectral radiance values scaled to minimum and maximum quantized calibrated pixel values. The product of certain gain is then multiplied with the QCAL value of pixel summing up with a bias, which is no other than spectral radiance scaled to minimum QCAL value. QCAL value range depends on the Radiometric Resolution of the sensor. In this work, dataset of AWiFS images is used which is of 12 bit radiometric resolution i.e. 0(MIN) to 4095(MAX). 

Spectral radiance is defined by the symbol 'L$\lambda$':
\small
\begin{equation}
L\lambda = gain * QCAL + bias
\end{equation}\label{formula:Llambda}
\begin{equation}
    gain = \frac{LMAX\lambda - LMIN\lambda}{QCALMAX - QCALMIN}
\end{equation}
where:
\begin{conditions}
$bias$ & LMIN$\lambda$\\
$QCAL$ & quantized calibrated pixel value in Digital Number(DN).\\
$LMIN \lambda$ & spectral radiance scaled to QCALMIN in $watts/(meter squared * steradian * \mu m)$.\\
$LMAX \lambda$ & spectral radiance scaled to QCALMAX in $watts/(meter squared * steradian * \mu m)$.\\
$QCALMIN$ & minimum quantized calibrated pixel value (corresponding to LMIN$\lambda$) in DN\\
 & 1 for LPGS products\\
 & 1 for NLAPS products processed after 4/4/2004 \\
 & 0 for NLAPS products processed before 4/5/2004 \\
$QCALMAX$ & maximum quantized calibrated pixel value (corresponding to LMAX$\lambda$) in DN.\\
 & 4095 for 12 bit.`2'
\end{conditions}
\normalsize
\subsection{TOA Reflectance}

The Top-Of-Atmosphere (TOA) reflectance calculation is an important preprocessing step in satellite image processing. The calibrated DN values of pixel should be corrected before application of image processing algorithm. While capturing images from satellite sensor, the earth's surface as well as atmospheric entities reflect energy (Irradiance) which affects the visibility of sensor. These entities cause exo-atmospheric irradiation, which is mainly solar irradiation \cite{chavez1996image}. Equation \ref{formula:TOAreflectance} is used \cite{chavez1996image} to convert the satellite radiance values into reflectance using exo-atmospheric irradiation, spectral radiance at the sensor’s aperture, sun-to-earth distance correction and solar zenith angle. It is remarkable that the calculation of solar zenith angle requires precise longitude and latitude values. The use of TOA reflectance for the image processing has the advantage that no other correction/conversion (e.g. atmospheric correction, angular effects and topographic (illumination) corrections) is needed. Moreover, the actual radiometric measurements can be used for retrieval of desired pixels \cite{subramaniam2011automated}.

TOA reflectance is denoted by $\rho$TOA \cite{TOAReflectance} and is computed as follows:
\small
\begin{equation} \label{formula:TOAreflectance}
\rho TOA = \frac{\pi.L\lambda.d^2}{ESUN.cos(\theta_S)}
\end{equation}
where:
\begin{conditions}
$\rho$TOA & TOA reflectance. \\
$L \lambda$ & spectral radiance at sensor aperture (mWcm$^{-2}$sr$^{-1}$$\mu$).\\
$d$ & Earth–Sun distance (Astronomical Units). \\
$ESUN$ & mean solar-exoatmospheric irradiance (mWcm$^{-2}$sr$^{-1}$$\mu$). \\
$\theta_S$ & solar zenith angle (degrees).
\end{conditions}
\normalsize
\section{Distinguishing Features}
\label{choiceOfFeatures}


\subsection{Normalized Difference Vegetation Index (NDVI)}

The satellite image processing methods have some fundamental methods to figure out the various land covers like agricultural region, non-agricultural region, forests, grasslands, etc. To perform such classification and to detect the various regions, the normalized differences between the separate reflectance values of pixels are necessary. These values are generally known as Normalized Difference Vegetation Indexes (NDVI). The cluster of these vegetation values can be used to visualize graphically and analyze the remote sensing measurements from a space platform, but not necessarily always. It can also used to assess whether the area of interest contains the live green vegetation or not \cite{gerard2003forest}. Mathematically, the summation and difference of the two spectral bands contains the same information as the original data, however the difference alone (or the normalized difference) carries only a part of the initial information. Sometimes, the missing information is relevant for the system to observe the whole scenario. However, it is important to understand that the NDVI product carries only a fraction of the information available in the original spectral reflectance data of source.
\small
\begin{equation}
NDVI = \frac{b3 - b4}{b3 + b4}
\end{equation}
\normalsize
where, b3 and b4 are the reflectance values of red and NIR band images at a specific pixel location.

Many systems have used Normalized Difference Water Index (NDWI) and Modified Normalized Difference Water Index (MNDWI) in their algorithm to find out their required results like water bodies in different forms. Although, these are promising values but they also comes with ambiguity between muddy water and cloud shadows. Also, there is a very fine difference in the values of the real burnt area and muddy water too. In this work we have obtained the results solely based on NDVI and BRT. 

\subsection{Brightness}

Brightness is the summation of the TOA reflectance values of all the band images from AWiFS dataset. The bands included in this calculation are B2 (Green), B3 (Red), B4 (NIR), B5 (SWIR) from AWiFS of Resourcesat-2 satellite. According to the following formula, the each B2, B3, B4, B5 are the values from same index of all of the bands respectively. Hence the resultant value should be mapped on same index as sources \cite{subramaniam2011automated}.
\small
\begin{equation}
Brightness (BRT) = B2 + B3 + B4 + B5
\end{equation}
\normalsize
The brightness values should be calculated for each pixel and hence the result will be the single two-dimensional matrix of same dimension as source images. This should be consistent because the different areas show distant values of brightness like water, soil, chlorophyll, ash, fire flares, cloud, etc. Their respective location will be the prime part in their detection, extraction and classification from other types of vegetation areas. On the basis of this, the brightness could be used as one of the distinguishing feature in algorithm of differentiating the water pixels from non-water pixels. 

\subsection{Sparse Feature Vector}
Fundamentally, it is critical to store the chosen features discussed in section \ref{choiceOfFeatures}. It is vital that information about the water bodies and burnt areas from the source image is not lost, as it is required in the future for further statistical and analytical tasks like computing similarity, training machine learning models, etc. The computer vision community has been concerned about storing such information for a long time and solve this by exploiting sparseness in the data.\cite{agarwal2002learning} \cite{henaff2011unsupervised} \cite{huang2006learning}.
Here, we devise a Sparse Feature Vector (SFV) consisting of the following features after preprocessing,\textbf{ (index(x, y), b2, b3, b4, b5, NDVI, Brightness)}. It is remarkable to note that the SFV intelligently stores only the important information, i.e the index and influencing features, thereby saving tons of spatial and computational resources.

\begin{wrapfigure}{R}{0.2\textwidth}
\centering
\includegraphics[width=0.2\textwidth]{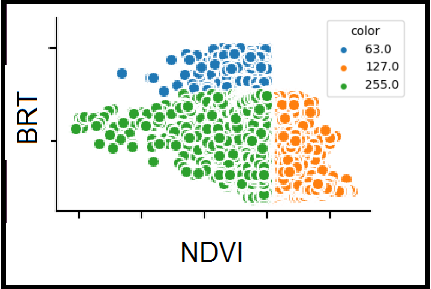}
\caption{Distinguishing Features}
\label{fig:DistinguishingFeatures}
\end{wrapfigure}

Table \ref{SparseFeatures} shows that, on an average, the SFV generally saves around 90\% - 95\% computational resources on a 2264 x 2264 resolution satellite image. Each row in Table \ref{SparseFeatures} corresponds to a Tile Number (e.g. NH44N), the date at which it was captured and the number of water pixels from our Dynamic Semantic Segmentation Algorithm (see Section \ref{DSS}) and percentage of relevant pixels (ones that contribute to the feature vector).

Figure \ref{fig:DistinguishingFeatures} represents clusters of three output classes of water pixels: 255.0 (Clear), 127.0 (Moderately Clear) and 63.0 (Muddy). It is evident that these classes form distinct clusters and can serve as ground truth labels for training deep models.

Hence, it is clear that most satellite images are highly sparse in nature and need not be completely stored but only a fraction of them is of relevance. The SFV is featured engineered in such a way that it not only generalizes well for Water Bodies but also for Burnt Areas. However, additional feature \textbf{BAIM} (see Section \ref{burntarea}) is required for finding Burnt pixels. Since, we have already computed water and burnt area percentages using the algorithm in Section \ref{DSS} and Sparse Feature Vector stores the indices, we can easily compute the Similarity by coupling percentages and mapping indices of the reference image to the ones in the database.

{\renewcommand{\arraystretch}{1.0}
\begin{table}[ht]
\caption{Reduction in Resource Utilization using Sparse Feature Vector (SFV)}
\centering
\begin{tabular}{ |c | c| c | c | } 
\hline
\multicolumn{1}{{|>{\centering\arraybackslash}m{13mm}|}}{\textbf{ Tile Number}} & \multicolumn{1}{{|>{\centering\arraybackslash}m{25mm}|}}{\textbf{Captured Date}} & \multicolumn{1}{{|>{\centering\arraybackslash}m{20mm}|}}{\textbf{Relevant Pixels}} \\ [1.0ex]
&  & \textbf{(in \%)}\\
\hline \hline
NH44N & \formatdate{9}{5}{2013} & 0.1\\ 
\hline
NH44N & \formatdate{19}{4}{2012} & 0.7\\
\hline
NH44N & \formatdate{25}{2}{2013}  & 5.9\\
\hline
NE43I & \formatdate{8}{1}{2013} & 1.0\\
\hline
NH44B & \formatdate{4}{9}{2012} & 3.8\\
\hline
NH44B & \formatdate{5}{8}{2013} & 3.99\\
\hline
NH44B & \formatdate{9}{5}{2013}  & 2.3\\
\hline
\end{tabular}
\label{SparseFeatures}
\end{table}
}
\section{Our Contribution - Dynamic Semantic Segmentation} \label{DSS}

The main contribution of this paper is two fold. We propose two Dynamic Semantic Segmentation (DSS) algorithms:
\begin{enumerate}
\item DSS for identifying multi-class water bodies at a pixel level
\item DSS for identifying burnt areas at a pixel level
\end{enumerate}
\begin{figure}[ht]
\centering
 \includegraphics[width=0.48\textwidth]{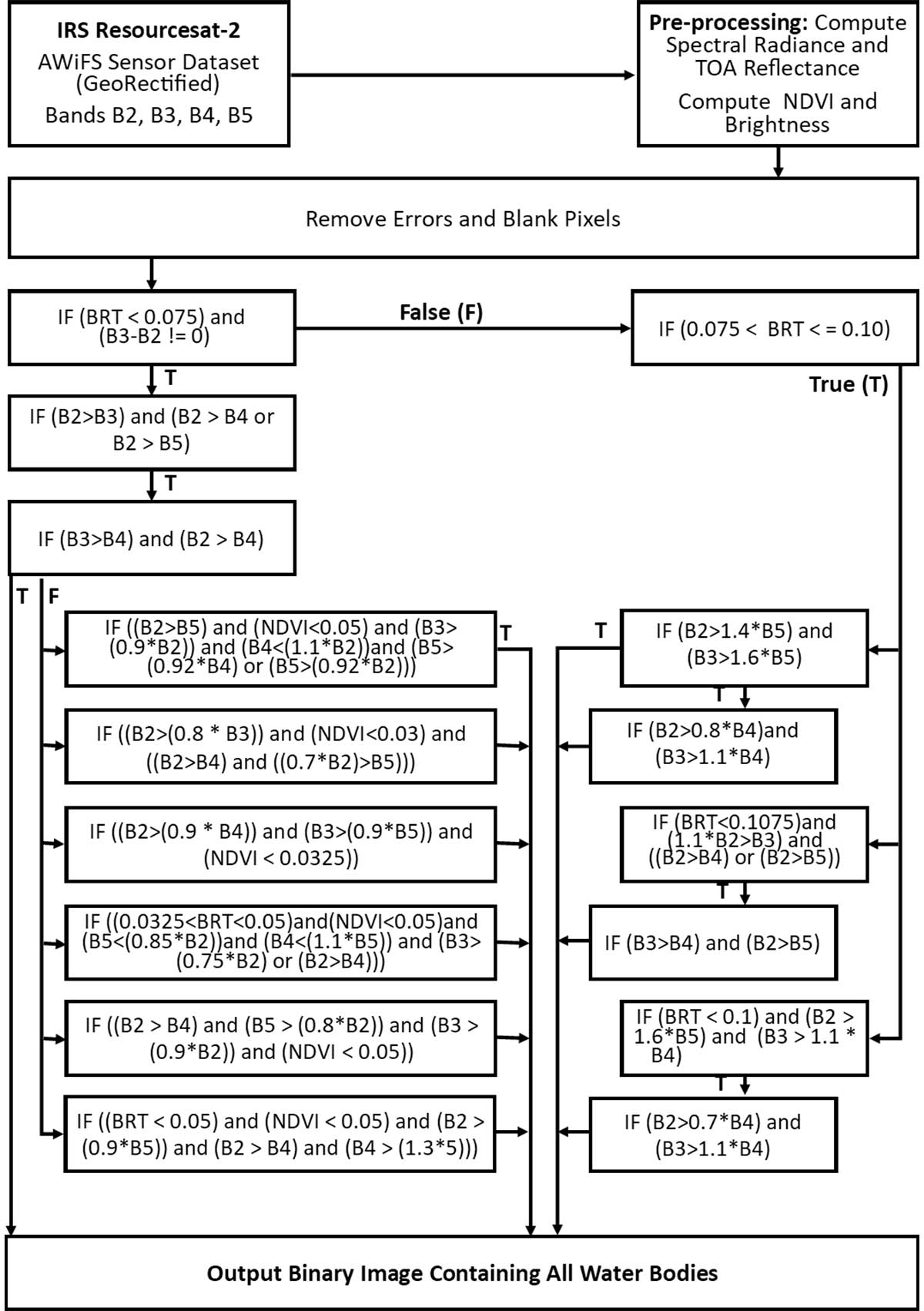}
 \caption{Dynamic Semantic Segmentation Algorithm for finding Water Bodies}
 \label{fig:DSSWater}
\end{figure}

\subsection{For Identifying Water Pixels} \label{waterbodies}
The algorithm (See figure \ref{fig:DSSWater}) operates sequentially at a pixel level and thereby for each pixel in the output image, it takes as input a correspondingly indexed pixel from each band. Since different types of water bodies (clear water, muddy water, bog, etc.) have distinct reflectance values (See figure \ref{fig:DistinguishingFeatures}), it is imperative to differentiate between them for multiple use cases. However, the distance between their representative pixels is minuscule. Our DSS algorithm filters out the irrelevant pixels in step by step fashion that will conserve the flow as well as protect the true representative pixels from filtration and mixing. The structure of the algorithm is like a decision tree with multiple levels wherein each level acts as a filter. 

\begin{figure}[ht]
\centering
 \includegraphics[width=0.48\textwidth]{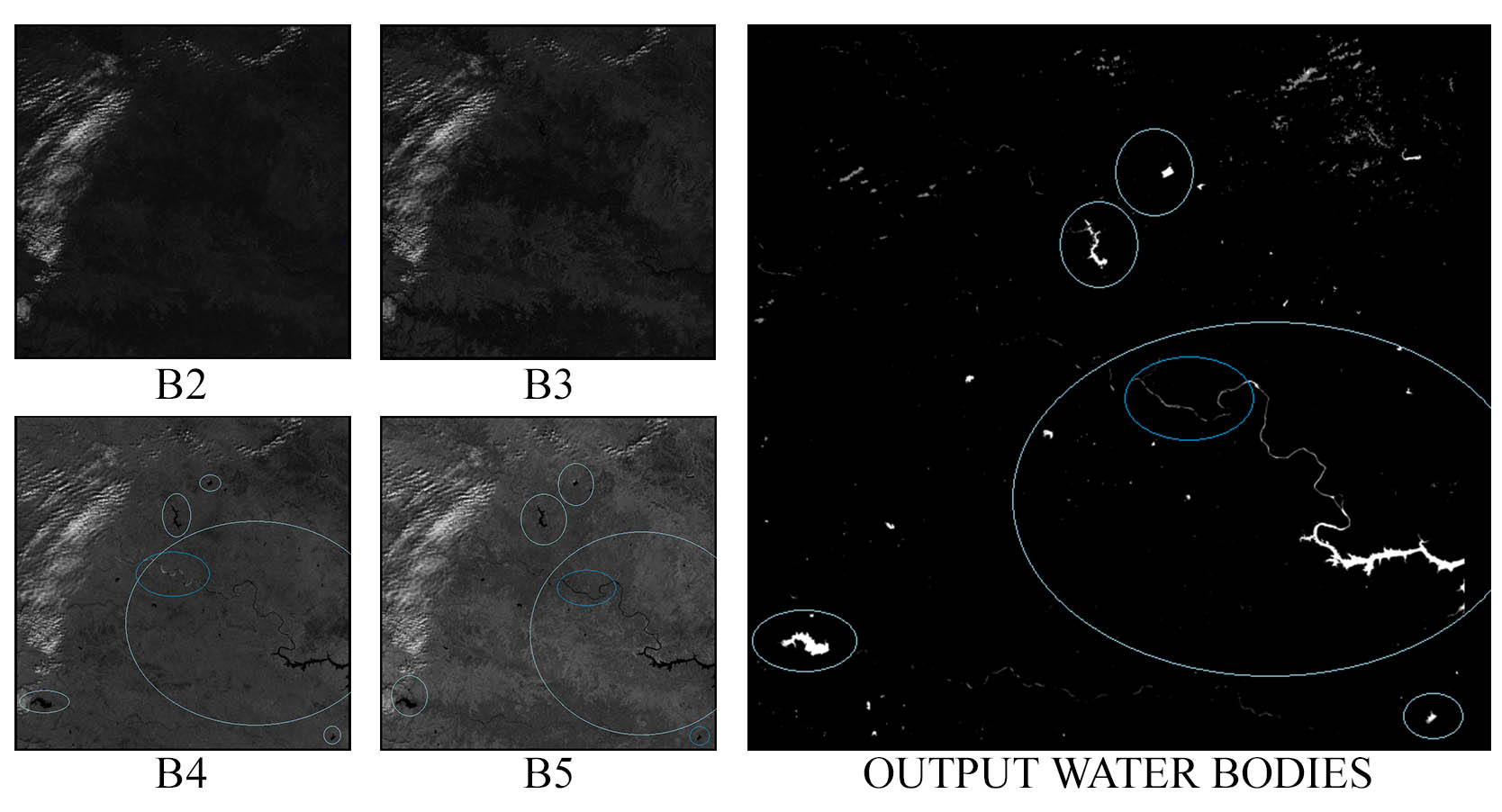}
 \caption{Results of DSS algorithm for water bodies}
 \label{fig:DSSWaterResults}
\end{figure}

Figure \ref{fig:DSSWaterResults} illustrates the results of our algorithm. Since water bodies have a predilection towards absorbing SWIR wavelengths, it makes them appear as dark bodies in the B5 (SWIR) band images (See figure \ref{fig:DSSWaterResults}). The ellipses in the input band images of B4 and B5 bands highlight visible water bodies. Evidently, these water bodies are from the same orthorectified tile and cannot be seen in the B2 and B3 bands. The output image clearly depicts the same water bodies at a very fine grained accuracy. Moreover, the pixels that appear to be grayish probably belong to a muddy water body while the brighter ones are from a clear water body.

\subsection{For Burnt Area Pixels} \label{burntarea}

Burnt area are significantly observed under specific range of values of each bands and can be obtained by applying empirically set thresholds of NDVI, BRT, and BAIM (Burnt Area Index by Martin) \cite{martin2005performance}. The BAIM threshold range was empirically obtained from AWiFS images having assured burnt areas/scars. From the analysis of the spectral values in Bands B4 and B5 (NIR and SWIR), the convergence values were fixed at 0.05(NIR) and 0.02(SWIR). These values are not the expected for severe burnings, but for smaller ones in B4 and around the average values of B5. We modified the convergence value according to formula \ref{formula:BAIM} to avoid confusion with other land covers, which are spectrally close to burnt area, such as muddy water. It should be kept in mind that burned area includes a wide range of fire severities, and therefore, a wide variety of mixtures between charcoal, ash, soil and scorched vegetation. The BAIM was intended to discriminate between burned and unburned areas, instead of discriminating different burn severities.

Burnt Area Index is given by
\small
\begin{equation} \label{formula:BAIM}
BAIM = \frac{1}{(pc_{NIR} - \rho_{NIR})^2 + (pc_{SWIR} - \rho_{SWIR})^2}
\end{equation}
where:
\begin{conditions}
$pc_{NIR}$ & convergence value for NIR channel. Constant value (0.05).\\
$pc_{SWIR}$ & convergence value for SWIR channel. Constant value (0.02).\\
$\rho_{NIR}$ & TOA value for NIR channel.\\
$\rho_{SWIR}$ & TOA value for SWIR channel.
\end{conditions}
\normalsize

Algorithm \ref{algo:probableburnt} contains a very acute range of those values which are obtained after rigorous experimentation on multiple tiles. Fundamentally, the algorithm contains two sets of variables. The first set acts like a filter and uses input pixels from each of the four bands for thresholding. However, it was empirically observed that this filter was not generic enough to differentiate between many corner cases.

\begin{algorithm}
\SetAlgoLined
\KwResult{Probable Burnt Pixels }
\caption{Finding Probable Burnt Area Pixels}
\label{algo:probableburnt}
 preprocessing\;
 \eIf{(0.0085 $<$ B2 $<$ 0.010) and (0.004 $<$ B3 $<$ 0.008) and (0.0085 $<$ B4 $<$ 0.0170) and (0.01 $<$ B5 $<$ 0.0130)}{
    probably a burnt pixel\;
   }{
   probably not  a burnt pixel\;
   }
\end{algorithm}

Hence, information from the input pixels wasn't sufficient for confidently classifying a pixel into a burnt pixel. Therefore, additional features: Brightness, NDVI and BAIM were tuned and added to the algorithm as a secondary check. The second set sieves out the remaining unburnt pixels. (See Algorithm \ref{algo:surelyburnt}).
$\newline$
$\newline$

\begin{algorithm}
\SetAlgoLined
\KwResult{Sure Burnt Pixels }
\caption{Finding Sure Burnt Area Pixels}
\label{algo:surelyburnt}
 preprocessing\;
 \eIf{((0.0085 $<$ B2 $<$ 0.010) and (0.004 $<$ B3 $<$ 0.008) and (0.0085 $<$ B4 $<$ 0.0170) and (0.01 $<$ B5 $<$ 0.0130))}{
 	\eIf{((585.0 $<$= BAIM $<$= 925.0) and (0.09 $<$= NDVI $<$= 0.45) and(0.035 $<$= BRT $<$= 0.05)) }{
    Surely a burnt pixel\;
    }
    {}
    }
  {
    Surely not a burnt pixel\;
  }
\end{algorithm}

\begin{figure}[ht]
\centering
 \includegraphics[width=0.4\textwidth]{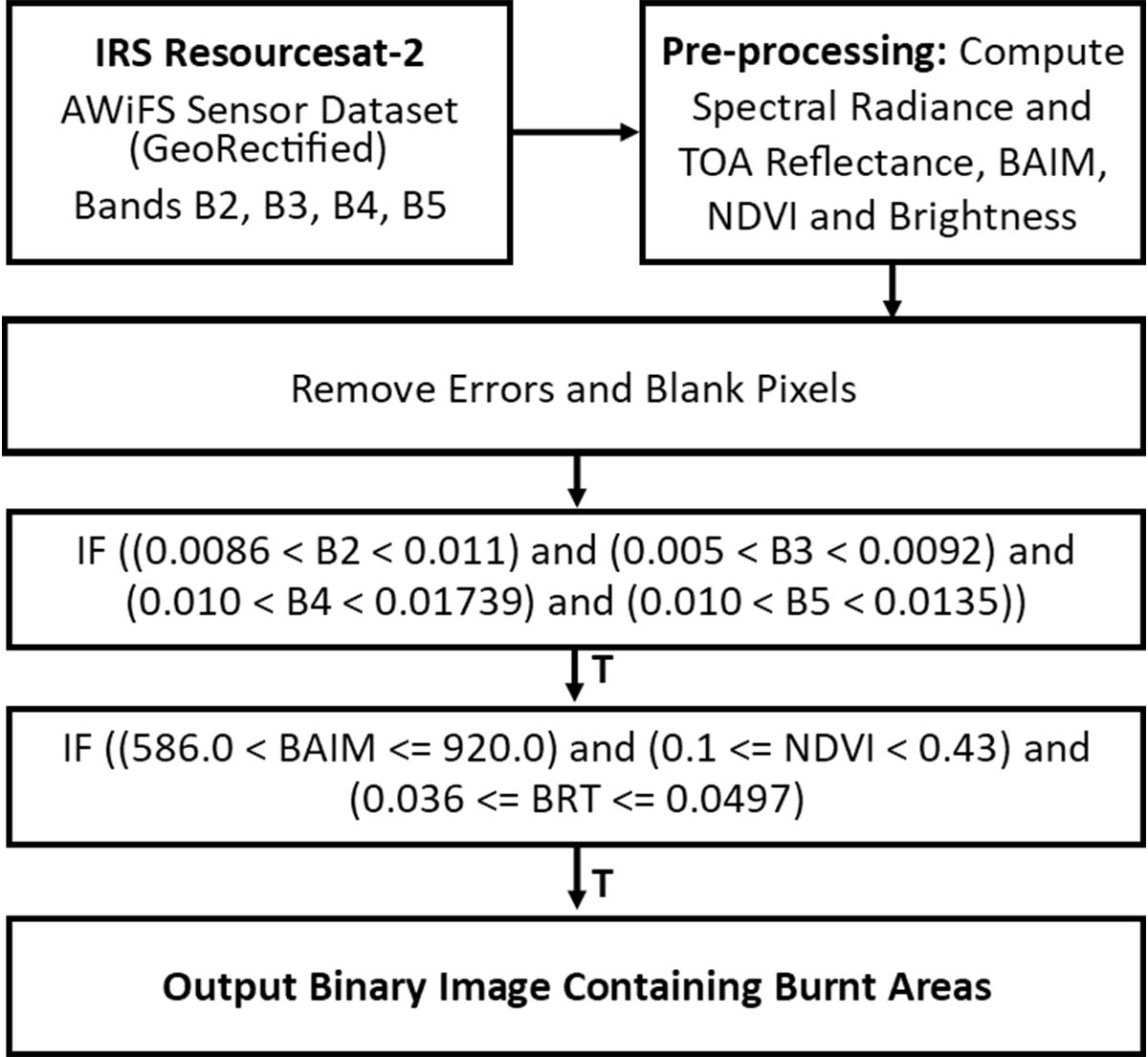}
 \caption{Dynamic Semantic Segmentation Algorithm for finding Burnt Area}
 \label{fig:DSSBurnt}
\end{figure}

\begin{figure}
\centering
 \includegraphics[width=0.48\textwidth]{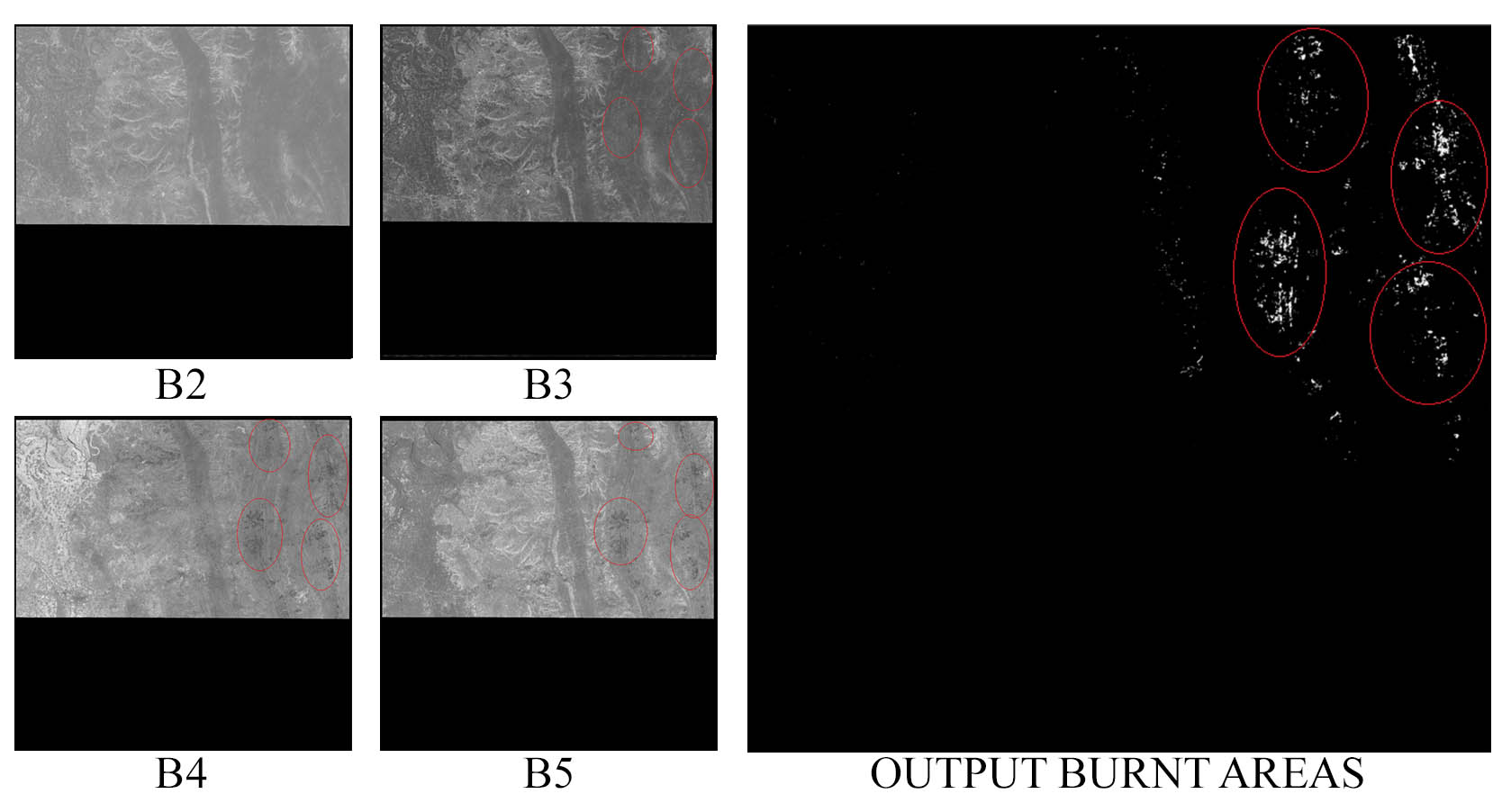}
 \caption{Results of DSS algorithm for burnt areas}
 \label{fig:DSSBurntResults}
\end{figure}

Analogous to the algorithm in figure \ref{fig:DSSWater}, DSS algorithm for finding burnt areas by operating sequentially at a pixel level as it processes corresponding pixel from each band. However, burnt areas are not classified further in the hierarchy. Figure \ref{fig:DSSBurntResults} illustrates the results of our algorithm \ref{algo:surelyburnt}. The burnt areas are slightly visible to the naked eye in B3, B4, B5 (black regions inside marked ellipses). Even though B2 is the same orthorectified tile, it is not possible to see burnt areas because burnt scars are readily visible in NIR Band image \cite{sedano2012increasing}. The output image lucidly portrays a binary mask of the same tile with white pixels denoting burnt areas.

\section{Observations and Conclusions}

There are some parameters which must be taken under consideration while processing AWiFS images with different radiometric resolution. For instance, the QCAL range (see equation \ref{formula:Llambda}) should be adjusted according to radiometric resolution.

Based on multiple experiments, our algorithm (see figure \ref{fig:DSSWater}) successfully segments water bodies and furthermore distinguishes clear water from muddy water and marsh areas. It was experimentally found that the misclassification of cloud shadows as muddy water by using algorithm by Subramaniam et al. (see Section \ref{Current State of Art}) can be further removed using our algorithm (see figure \ref{fig:DSSWaterResults}).

The output images are used to calculate the percentage of water bodies using equation \ref{waterbodypercnteqn}. These values stored along with tile information have multiple use cases in geographical surveys and analysis.
\begin{equation} \label{waterbodypercnteqn}
Water Bodies (\%) = \frac{Water Pixels}{TotalPixels} * 100
\end{equation}

Similarly, another algorithm defined in this paper (see figure \ref{fig:DSSBurnt}) has proved to be helpful in detection of burnt areas. Burnt area detection using AWiFS alone is found to be rare. Hence, there is not enough research in this area. Most of the state-of-art research work in this domain depends on multiple datasets of images to corner the burnt area (see section \ref{Current State of Art}). However, this work requires reflectance values from different bands of AWiFS satellite images and does not have any other dependencies. Moreover, dependencies on data from sensors like Moderate Resolution Imaging Spectroradiometer (MODIS) which use thermal bands for finding burnt area is also eliminated.

Similar to equation \ref{waterbodypercnteqn}, burnt area output images are used to calculate the percentage of burnt areas in respective tile image using the equation \ref{burntareapercnteqn}. These values can be stored in database for future applications.
\begin{equation} \label{burntareapercnteqn}
Burnt Area (\%) = \frac{Burnt Pixels}{TotalPixels} * 100
\end{equation}

The above computed percentages and the sparse feature vectors stored in database can be used to compute similarity. Overall, the system can be deployed in various use cases like analysis of flood, drought conditions, forest fire management, damage assessment, etc. The database schema is designed as such to enable query based retrieval, for example, retrieval of images based on water content or burnt area, retrieval of images based on location, etc. Using orthorectified tiles along with their tile IDs (region of earth's surface) and the sparse feature vectors stored in database, this work gives an optimized way find similar images and rank them according to the percentage of water bodies and burnt areas found in them.

Table \ref{table:ReflectanceRanges} demonstrates the reflectance range values precise to 4 decimal places obtained from the proposed DSS algorithms.

{\renewcommand{\arraystretch}{1.2}
\begin{table}
\caption{Range of possible reflectance values for each feature empirically obtained from Dynamic Semantic Segmentation (DSS) algorithm for pixels belonging to a clear water body, muddy water body or burnt area.}
\centering
\resizebox{\columnwidth}{!}{%
\begin{tabular}{|c|c|c|c|}
\hline
\textbf{ Features } & \textbf{ Clear water } & \textbf{ Muddy water } & \textbf{ Burnt } \\
\hline
B2 & 0.0078 --- 0.0142 & 0.0076 --- 0.0239 & 0.0092 --- 0.0100 \\
\hline
B3 & 0.0046 --- 0.0118 & 0.0048 --- 0.0233 & 0.0071 --- 0.0084 \\
\hline
B4 & 0.0047 --- 0.0228 & 0.0159 --- 0.0285 & 0.0092 --- 0.0154 \\
\hline
B5 & 0.0037 --- 0.0221 & 0.0148 --- 0.0276 & 0.0087 --- 0.0155 \\
\hline
BRT & 0.0255 --- 0.0677 & 0.0475 --- 0.0992 & 0.0343 --- 0.0479 \\
\hline
NDVI & -0.1746 --- 0.6600 & -0.0492 --- 0.608 & 0.1203 --- 0.3598 \\
\hline
BAIM & - & - & 558.3829 --- 922.4521 \\
\hline
\end{tabular}\label{table:ReflectanceRanges}
}
\end{table}
}
\bibliographystyle{ieee}
\bibliography{main.bib}

\begin{thebibliography}{10}\itemsep=-1pt

\bibitem{agarwal2002learning}
S.~Agarwal and D.~Roth.
\newblock Learning a sparse representation for object detection.
\newblock In {\em European conference on computer vision}, pages 113--127.
  Springer, 2002.

\bibitem{chavez1996image}
P.~S. Chavez et~al.
\newblock Image-based atmospheric corrections-revisited and improved.
\newblock {\em Photogrammetric engineering and remote sensing},
  62(9):1025--1035, 1996.

\bibitem{gerard2003forest}
F.~Gerard, S.~Plummer, R.~Wadsworth, A.~F. Sanfeliu, L.~Iliffe, H.~Balzter, and
  B.~Wyatt.
\newblock Forest fire scar detection in the boreal forest with multitemporal
  spot-vegetation data.
\newblock {\em IEEE Transactions on Geoscience and Remote Sensing},
  41(11):2575--2585, 2003.

\bibitem{henaff2011unsupervised}
M.~Henaff, K.~Jarrett, K.~Kavukcuoglu, and Y.~LeCun.
\newblock Unsupervised learning of sparse features for scalable audio
  classification.
\newblock In {\em ISMIR}, volume~11, page 2011. Citeseer, 2011.

\bibitem{huang2006learning}
C.~Huang, H.~Ai, Y.~Li, and S.~Lao.
\newblock Learning sparse features in granular space for multi-view face
  detection.
\newblock In {\em Automatic Face and Gesture Recognition, 2006. FGR 2006. 7th
  International Conference on}, pages 401--406. IEEE, 2006.

\bibitem{ISRO}
ISRO.
\newblock {\em ISRO}, 2012.

\bibitem{Resourcesat-2_Handbook}
ISRO.
\newblock {\em Resourcesat-2_Handbook}, 2014.

\bibitem{martin2005performance}
M.~Martin, I.~Gomez, and E.~Chuvieco.
\newblock Performance of a burned-area index (baim) for mapping mediterranean
  burned scars from modis data.
\newblock In {\em Proceedings of the 5th international workshop on remote
  sensing and GIS applications to forest fire management: fire effects
  assessment}, pages 193--198. Paris, Universidad de Zaragoza, GOFC GOLD,
  EARSeL, 2005.

\bibitem{BHUVAN}
NRSC.
\newblock {\em Bhuvam Web Portal}, 2012.

\bibitem{sedano2012increasing}
F.~Sedano, P.~Kempeneers, P.~Strobl, D.~McInerney, and J.~San~Miguel.
\newblock Increasing spatial detail of burned scar maps using irs-awifs data
  for mediterranean europe.
\newblock {\em Remote Sensing}, 4(3):726--744, 2012.

\bibitem{subramaniam2011automated}
S.~Subramaniam, A.~S. Babu, and P.~S. Roy.
\newblock Automated water spread mapping using resourcesat-1 awifs data for
  water bodies information system.
\newblock {\em IEEE Journal of Selected Topics in Applied Earth Observations
  and Remote Sensing}, 4(1):205--215, 2011.

\bibitem{TOAReflectance}
U.~S.~G. Survey.
\newblock {\em Overview of the Resourcesat-1 (IRS-P6)}, 2007.

\bibitem{RadianceUSGS}
U.~S.~G. Survey.
\newblock {\em How to calculate radiance}, 2018.

\bibitem{upreti2011content}
D.~Upreti.
\newblock {\em Content-based satellite cloud image retrieval}.
\newblock University of Twente Faculty of Geo-Information and Earth Observation
  (ITC), 2011.

\end{thebibliography}

\end{document}